\pdfoutput=1
\documentclass[11pt]{article}

\usepackage{ACL2023}
\usepackage{times}
\usepackage{latexsym}
\usepackage{amsmath}
\usepackage{color}
\usepackage{booktabs}
\usepackage{graphicx}
\usepackage{subfigure}
\usepackage{ruby}
\usepackage{multirow}
\usepackage{courier}
\usepackage[T1]{fontenc}
\usepackage[utf8]{inputenc}
\usepackage{microtype}
\usepackage{inconsolata}
\usepackage{float}
\usepackage{overpic}
\usepackage{adjustbox}
\usepackage{mathptmx}
\usepackage{tcolorbox}
\usepackage{stackengine}

\makeatletter
\def\dual#1{\expandafter\dual@aux#1\@nil}
\def\dual@aux#1/#2\@nil{\begin{tabular}[t]{@{}c@{}}#1\\#2\end{tabular}}
\makeatother

\usepackage{listings}
\usepackage{color}
\usepackage{textcomp}
\definecolor{listinggray}{gray}{0.9}
\definecolor{lbcolor}{rgb}{0.9,0.9,0.9}
\newtcbox{\blue}[1][]{on line,boxsep=2pt,left=0pt,right=0pt,top=0pt,bottom=0pt,colframe=white,colback=blue!30!white,#1}
\newtcbox{\red}[1][]{on line,boxsep=2pt,left=0pt,right=0pt,top=0pt,bottom=0pt,colframe=white,colback=red!30!white,#1}
\newtcbox{\yellow}[1][]{on line,boxsep=2pt,left=0pt,right=0pt,top=0pt,bottom=0pt,colframe=white,colback=yellow!30!white,#1}
\newtcbox{\green}[1][]{on line,boxsep=2pt,left=0pt,right=0pt,top=0pt,bottom=0pt,colframe=white,colback=green!30!white,#1}
\newtcbox{\orange}[1][]{on line,boxsep=2pt,left=0pt,right=0pt,top=0pt,bottom=0pt,colframe=white,colback=orange!30!white,#1}

\newcommand{\transbox}[1]{\rotatebox{90}{#1}}

\title{On Evaluating Multilingual Compositional Generalization \\ with Translated Datasets}

\author{Zi Wang$^{1,2}$ \and Daniel Hershcovich$^{1}$ \\
  $^{1}$Department of Computer Science, \\$^{2}$Department of Nordic Studies and Linguistics\\University of Copenhagen\\
  \texttt{\{ziwa, dh\}@di.ku.dk}}

\begin{document}
\maketitle
\begin{abstract}

Compositional generalization allows efficient learning and human-like inductive biases. Since most research investigating compositional generalization in NLP is done on English, important questions remain underexplored. Do the necessary compositional generalization abilities differ across languages? Can models compositionally generalize cross-lingually? As a first step to answering these questions, recent work used neural machine translation to translate datasets for evaluating compositional generalization in semantic parsing. However, we show that this entails critical semantic distortion. To address this limitation, we craft a faithful rule-based translation of the MCWQ dataset \cite{cui-etal-2022-compositional} from English to Chinese and Japanese. Even with the resulting robust benchmark, which we call MCWQ-R, we show that the distribution of compositions still suffers due to linguistic divergences, and that multilingual models still struggle with cross-lingual compositional generalization. Our dataset and methodology will be useful resources for the study of cross-lingual compositional generalization in other tasks.\footnote{The dataset, trained models and code for the experiments and dataset generation are available at \url{https://github.com/ziwang-klvk/CFQ-RBMT}.}

\end{abstract}

\section{Introduction}\label{intro}
A vital ability desired for language models is compositional generalization (CG), the ability to generalize to novel combinations of familiar units \citep{oren-etal-2020-improving}. Semantic parsing enables executable representation of natural language utterances for knowledge base question answering \citep[KBQA;][]{KBQAsurvey}. A growing amount of research has been investigating the CG ability of semantic parsers based on carefully constructed datasets, typically synthetic corpora \citep[e.g., CFQ;][]{CFQ} generated based on curated rules, mostly within monolingual English scenarios. As demonstrated by \citet{multilingKBQA}, resource scarcity for many languages largely preclude their speakers' access to knowledge bases (even for languages they include), and KBQA in multilingual scenarios is barely researched mainly due to lack of corresponding benchmarks.

\begin{CJK*}{UTF8}{gbsn}
\begin{figure}[t]
\footnotesize{\textsc{Neural-based Translation:}} \\
{\scriptsize
\textsc{Source: }\textit{Did\red{Erika Mann's spouse executive}\red{produce}\blue{Friedemann Bach}}\\
\textsc{Target: }\red{艾莉卡·曼的配偶执行官}\red{制作}了\blue{弗里德曼·巴赫}吗
}

\vspace{4mm}
\footnotesize{\textsc{Rule-based Translation:}} \\
{\scriptsize
\textsc{Source: }\textit{Did\orange{Erika Mann's spouse}\green{executive produce}\blue{Friedemann Bach}\\
\textsc{Target: }\dual{\orange{艾莉卡·曼的配偶}/\orange{$_{\textsc{NP1}}$}}\dual{\green{执行制作}/\green{$_{\textsc{V}}$}}了\dual{\blue{弗里德曼·巴赫}/\blue{$_{\textsc{NP2}}$}}吗}

\vspace{4mm}
\footnotesize{\textsc{SPARQL Query:}}\\
{\tt \scriptsize
\textbf{ASK WHERE} \{ 
\blue{wd:Q829979}  \green{wdt:P1431}  \orange{?x0} . \\ \orange{?x0  wdt:P26 wd:Q61597} . FILTER ( \orange{?x0} != wd:Q61597 )\}
}}
    \caption{Example of neural machine translation (NMT, from MCWQ, top) and rule-based translation (from MCWQ-R, middle) from English to Chinese. The compositions correctly captured by the translation system and the correspondences in the SPARQL query (bottom) are highlighted in the same color, while errors are in red. NMT often diverges semantically from the query: here, the compound ``executive produce'' is split. RBMT performs well due to awareness of grammar constituents.}
    \label{fig:preface}
\end{figure}
\end{CJK*}

\citet{cui-etal-2022-compositional} proposed Multilingual Compositional Wikidata Questions (MCWQ) as the first semantic parsing benchmark to address the mentioned gaps. Google Translate \citep[GT;][]{GT}, a Neural Machine Translation (NMT) system trained on large-scale corpora, was adopted in creating MCWQ. We argue that meaning preservation during translation is vulnerable in this methodology especially considering the synthetic nature of the compositional dataset. Furthermore, state-of-the-art neural network models fail to capture structural systematicity \citep{hadley1994systematicity, SCAN, kim-linzen-2020-cogs}. 

Symbolic (e.g., rule-based) methodologies allow directly handling CG and were applied both to generate benchmarks \citep{CFQ, kim-linzen-2020-cogs, tsarkov2021starcfq} and to inject inductive bias to state-of-the-art models \citep{HPD, liu-etal-2021-learning-algebraic}. This motivates us to extend this idea to cross-lingual transfer of benchmarks and models. We propose to utilize rule-based machine translation (RBMT) to create parallel versions of MCWQ and yield a robust multilingual benchmark measuring CG. We build an MT framework based on synchronous context-free grammars (SCFG) and create new Chinese and Japanese translations of MCWQ questions, which we call MCWQ-R (Multilingual Compositional Wikidata Questions with Rule-based translations). We conduct experiments on the datasets translated with GT and RBMT to investigate the effect of translation method and quality on CG in multilingual and cross-lingual scenarios.

Our specific contributions are as follows:
\begin{itemize}
    \item We propose a rule-based method to faithfully and robustly translate CG benchmarks.
    \item We introduce MCWQ-R, a CG benchmark for semantic parsing from Chinese and Japanese to SPARQL.
    \item We evaluate the translated dataset through both automatic and human evaluation and show that its quality greatly surpasses that of MCWQ \cite{cui-etal-2022-compositional}.
    \item We experiment with two different semantic parsing architectures and provide an analysis of their CG abilities within language and across languages.
\end{itemize}

\section{Related Work}\label{background}

\paragraph{Compositional generalization benchmarks.}
Much previous work on CG investigated how to measure the compositional ability of semantic parsers. \citet{SCAN} and \citet{bastings-etal-2018-jump} evaluated the CG ability of sequence-to-sequence (seq2seq) architectures on natural language command and action pairs. \citet{CFQ} brought this task to a realistic scenario of KBQA by creating a synthetic dataset of questions and SPARQL queries, CFQ, and further quantified the distribution gap between training and evaluation using \textit{compound divergence}, creating maximum compound divergence (MCD) splits to evaluate CG. Similarly, \citet{kim-linzen-2020-cogs} created COGS in a synthetic fashion following a stronger definition of training-test distribution gap.
\citet{goodwin-etal-2022-compositional} benchmarked CG in dependency parsing by introducing gold dependency trees for CFQ questions. For this purpose, a full coverage context-free grammar over CFQ was constructed benefiting from the synthetic nature of the dataset.
While these works differ in data generation and splitting strategy, rule-based approaches are commonly adopted for dataset generation; as \citet{kim-linzen-2020-cogs} put it, such approaches allow maintaining ``full control over the distribution of inputs'', the crucial factor for valid compositionality measurement.
In contrast, \citet{cui-etal-2022-compositional} created MCWQ through a process including knowledge base migration and question translation through NMT, without full control over target language composition distribution. We aim to remedy this in our paper by using RBMT.


\paragraph{Rule-based machine translation.}
Over decades of development, various methodologies and technologies were introduced for the task of Machine Translation (MT). To roughly categorize the most popular models, we can divide them into pre-neural models and neural-based models. Pre-neural MT \citep{wu-1996-polynomial, marcu-wong-2002-phrase, koehn-etal-2003-statistical, chiang-2005-hierarchical} typically includes manipulation of syntax and phrases, whereas neural-based MT \citep{kalchbrenner-blunsom-2013-recurrent, cho-etal-2014-learning, Transformer} refers to those employing neural networks. However, oriented to general broad-coverage applications, most models rely on learned statistical estimates, even for the pre-neural models.
The desiderata in our work, on the other hand, exclude methods with inherent uncertainty. The most relevant methods were by \citet{wu-1996-polynomial, wu-1997-stochastic} who applied SCFG variants to MT \citep{intro2SCFG}. The SCFG is a generalization of CFG (context-free grammars) generating coupled strings instead of single ones, exploited by pre-neural MT works for complex syntactic reordering during translation. In this work, we exclude the statistical component and manually build the SCFG transduction according to the synthetic nature of CFQ; we specifically call it ``rule-based'' instead of ``syntax-based'' to emphasize this subtle difference. 

\begin{figure*}[ht]
    \centering
    \includegraphics[width=\textwidth]{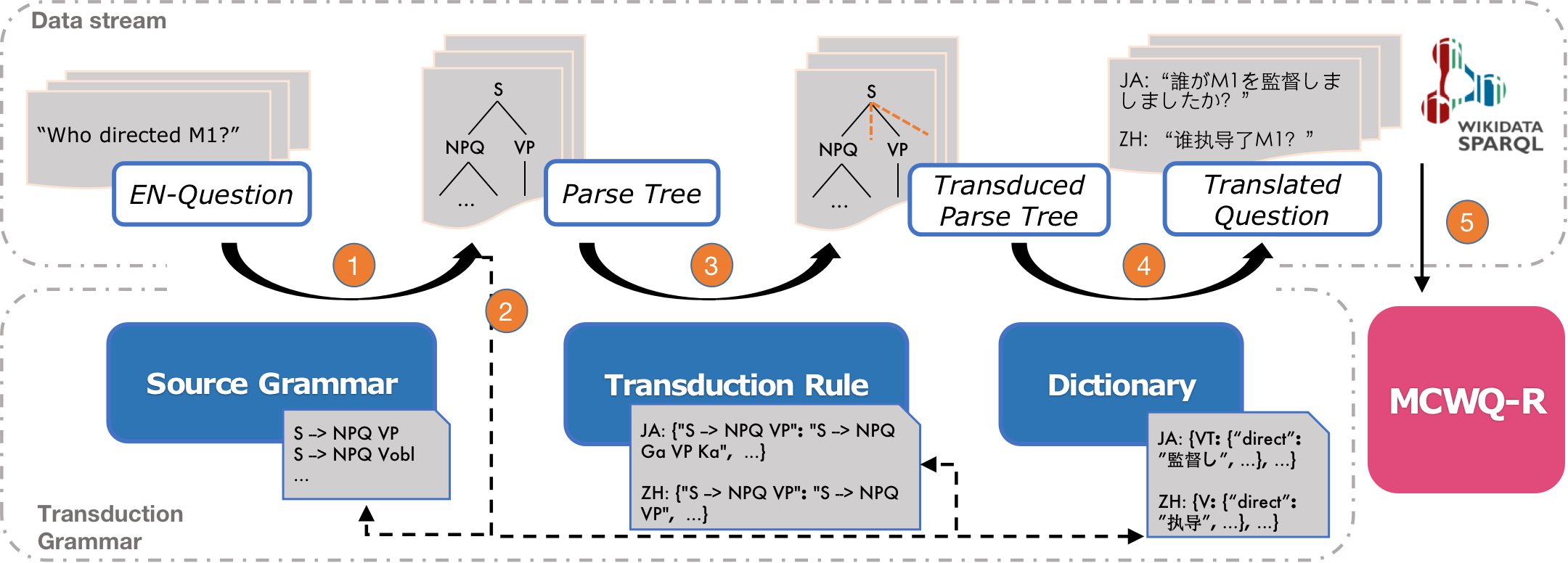}
    \caption{The pipeline of dataset generation. The circled numbers refer to (1) parsing question text, (2) building the dictionary and revising the source grammar and corresponding transduction rules based on parse trees, (3) replacing and reordering constituents, (4) translating lexical units, (5) post-processing and grounding in Wikidata.}
    \label{fig:pipeline}
\end{figure*}

\paragraph{Multilingual benchmarks.}
Cross-lingual learning has been increasingly researched recently, where popular technologies in NLP are generally adapted for representation learning over multiple languages \citep{conneau-etal-2020-unsupervised, xue-etal-2021-mt5}. Meanwhile, transfer learning is widely leveraged to overcome the data scarcity of low-resource languages \citep{cui-etal-2019-cross, hsu-etal-2019-zero}. However, cross-lingual benchmarks datasets, against which modeling research is developed, often suffer from ``translation artifacts'' when created using general machine translation systems \citep{artetxe-etal-2020-translation, wintner-2016-translationese}. \citet{longpre-etal-2021-mkqa} proposed MKQA, a large-scale multilingual question answering corpus (yet not for evaluating CG) avoiding this issue, through enormous human efforts. In contrast, \citet{cui-etal-2022-compositional} adopted Google Translate to obtain parallel versions for CFQ questions while sacrificing meaning preservation and systematicity. We propose a balance between the two methodologies, with automatic yet controlled translation. In addition, our work further fills the data scarcity gap in cross-lingual semantic parsing, being the first CG benchmark for semantic parsing for Japanese.


\section{Multilingual Compositional Wikidata Questions (MCWQ)}
MCWQ \citep{cui-etal-2022-compositional} is the basis of our work. It comprises English questions inherited from CFQ \citep{CFQ} and the translated Hebrew, Chinese and Kannada parallel questions based on Google Cloud Translate, an NMT system. The questions are associated with SPARQL queries against Wikidata, which were migrated from Freebase queries in CFQ. Wikidata is an open knowledge base where each item is allocated a unique, persistent identifier (QID).\footnote{\url{https://www.wikidata.org}}
MCWQ and CFQ (and in turn, our proposed MCWQ-R, see \S\ref{dataset}) share common English questions and associated SPARQL queries. MCWQ introduces distinct multilingual branches, with the same data size across all the branches.

\begin{CJK*}{UTF8}{gbsn}
\begin{figure*}
\begin{lstlisting}
"questionPatternModEntities": "Did M1 's spouse executive produce M0",
"questionWithBrackets": "Did [Erika Mann] 's spouse executive produce [Friedemann Bach]",
"questionPatternModEntities_zh": "M1 的配偶主管是否生产了 M0"
"questionWithBrackets_zh": "[Erika Mann] 的配偶执行官制作了 [Friedemann Bach] 吗",
"sparqlPatternModEntities": 
        "ASK WHERE { M0 wdt:P1431 ?x0 . ?x0 wdt:P26 M1 . FILTER ( ?x0 != M1 )}
"sparql": 
        "ASK WHERE { wd:Q829979 wdt:P1431 ?x0 . ?x0 wdt:P26 wd:Q61597 . FILTER ( ?x0 != wd:Q61597 )}"
\end{lstlisting}
\begin{lstlisting}[aboveskip=-0.1em]
"questionPatternModEntities_zh": "M1 的配偶执行制作了 M0 吗",
"questionWithBrackets_zh": "[Erika Mann] 的配偶执行制作了 [Friedemann Bach] 吗"
\end{lstlisting}
\caption{\label{fig:mcwq_example}
Example of an MCWQ \citep{cui-etal-2022-compositional} item in JSON format (top) and 2 fileds of the corresponding MCWQ-R item (bottom). We present part of the fields: the English and SPARQL fields inherited from CFQ and the Chinese fields. Specifically, we show an incorrectly translated example in MCWQ where ``excutive produce'' is not translated as a composition while MCWQ-R keeps good consistency with English.
}
\end{figure*}
\end{CJK*}
Due to the translation method employed in MCWQ, it suffers from detrimental inconsistencies for CG evaluation (see Figures \ref{fig:preface} and \ref{fig:mcwq_example})---mainly due to the unstable mapping from source to target languages performed by NMT models at both the lexical and structural levels. We discuss the consequences with respect to translation quality in \S\ref{quality} and model performance in \S\ref{analysis}.

\section{MCWQ-R: A Novel Translated Dataset}\label{dataset}
As stated in \S\ref{background}, data generation with GT disregards the ``control over distribution'', which is crucial for CG evaluation \citep{CFQ, kim-linzen-2020-cogs}. Thus, we propose to diverge from the MCWQ methodology by translating the dataset following novel grammar of the involved language pairs to guarantee controllability during translation. Such controllability ensures that the translations are deterministic and systematic. In this case, generalization is exclusively evaluated with respect to compositionality, avoiding other confounds.
We create new instances of MCWQ in Japanese and Chinese, two typologically distant languages from English, sharing one common language (Chinese) with the existing MCWQ. To make comprehensive experimental comparisons between languages, we also use GT to generate Japanese translations (which we also regard as a part of MCWQ in this paper), following the same method as MCWQ.

In this section, we describe the proposed MCWQ-R dataset.
In \S\ref{methodology} we describe the process of creating the dataset, in \S\ref{stats} its statistics, and in \S\ref{quality} the automatic and manual assessment of its quality.

\subsection{Generation Methodology}\label{methodology}
The whole process of the dataset generation is summarized in Figure \ref{fig:pipeline}.
We proceed by parsing the English questions, building bilingual dictionaries, a source grammar and transduction rules, replacing and reordering constituents, translating lexical units, post-processing and grounding in Wikidata.

\paragraph{Grammar-based transduction.}
We base our method on Universal Rule-Based Machine Translation \citep[URBANS;][]{URBANS}, an open-source toolkit\footnote{Released under the Apache 2.0 license: \url{https://github.com/pyurbans/urbans}.} supporting deterministic rule-based translation with a bilingual dictionary and grammar rule transduction, based on NLTK \cite{bird-loper-2004-nltk}. We modify it to a framework supporting synchronous context-free grammar \citep[SCFG;][]{intro2SCFG} for practical use, since the basic toolkit lacks \textit{links} from non-terminals to terminals preventing the lexical multi-mapping. A formally defined SCFG variant is symmetrical regarding both languages \citep{wu-1997-stochastic}, while we implement a simplified yet functionally identical version only for one-way transduction. Our formal grammar framework consists of three modules: a set of \textbf{source grammar} rules converting English sentences to parse trees, the associated \textbf{transduction rules} hierarchically reordering the grammar constituents with tree manipulation and a \textbf{tagged dictionary} mapping tokens into the target language based on their part-of-speech (POS) tags. The \textit{tagged} dictionary here provides \textit{links} between the non-terminals and terminals defined in a general CFG \citep{SBSMT}. Context information of higher syntactical levels is encapsulated in the POS tags and triggers different mappings to the target terms via the links. This mechanism enables our constructed grammar to largely address complex linguistic differences (polysemy and inflection for instance) as a general SCFG does. We construct the source grammar as well as associated transduction rules and dictionaries, resulting in two sets of transduction grammars for Japanese and Chinese respectively.

\paragraph{Source grammar.}
The synthetic nature of CFQ \citep{CFQ} indicates that it has limited sentence patterns and barely causes ambiguities; \citet{goodwin-etal-2022-compositional} leverage this feature and construct a full coverage CFG for the CFQ language, which provides us with a basis of source grammar. We revise this monolingual CFG to satisfy the necessity for translation with an ``extensive'' strategy, deriving new tags for constituents at the lowest syntactic level where the context accounts for multiple possible lexical mappings.

\paragraph{Bridging linguistic divergences.}
The linguistic differences are substantial between the source language and the target languages in our instances. The synthetic utterances in CFQ are generally \textit{cultural-invariant} and not entailed with specific language style, therefore the problems here are primarily ascribed to the grammatical differences and lexical gaps. For the former, our grammar performs systematic transduction on the syntactical structures; for the latter, we adopt a \textit{pattern match-substitution} strategy as post-processing for the lexical units applied in a different manner from the others in the target languages. We describe concrete examples in Appendix \ref{appendix:grammar}.
Without the confound of probability, the systematic transductions simply \textit{bridge} the linguistic gaps without further extension, i.e., no novel primitives and compositions are generated while the existing ones are faithfully maintained to the largest extent in this framework.

\paragraph{Grounding in Wikidata.}
Following CFQ and MCWQ, we ground the translated questions in Wikidata through their coupled SPARQL queries. Each \textit{entity} in the knowledge base possesses the unique QID and multilingual labels, meaning that numerous entities can be treated as simplified \textit{mod entities} (see Figure \ref{fig:mcwq_example}.) during translation, i.e., the grammar translates the \textit{question patterns} instead of concrete questions. The shared SPARQL queries enable comparative study with MCWQ and potentially CFQ (our grammar fully covers CFQ questions) in both cross-lingual and monolingual domains. 
In addition, the SPARQL queries are unified as reversible intermediate representation \citep[RIR;][]{rir} in our dataset and for all experimental settings, which is shown to improve CG.

\begin{table}[t]
\centering\small
\begin{tabular}{@{}ccclll@{}}
\toprule
& &  & Question  & Paired \\ 
& & Questions &  Patterns & Patterns \\ \midrule
\multicolumn{2}{c}{EN (MCWQ)}& 124,187 & 105,461 & 105,461 \\ \midrule
GT & JA & 124,187 & 99,900 & 100,140 \\
(MCWQ) & ZH & 124,187 & 99,747 & 100,325 \\ \midrule
RBMT & JA & 124,187 & 98,431 & 98,431 \\
(MCWQ-R) & ZH & 124,187 & 101,333 & 101,342 \\ \bottomrule
\end{tabular}
\caption{\label{stat}
Statistics of MCWQ-R and the corresponding branches of MCWQ. We count \textit{question patterns} with \textit{mod entities} \cite{CFQ}, the form directly processed during translation, and \textit{question-query pairs} comprising the question patterns and the associated SPARQL queries with mod entities.}
\end{table}

\subsection{Dataset Statistics}\label{stats}
Due to the shared source data, the statistics of MCWQ-R are largely kept consistent with MCWQ. Specifically, the two datasets have the same amounts of \textit{unique questions} (UQ; 124,187), \textit{unique queries} (101,856, 82\% of UQ) and \textit{query patterns} (86,353, 69.5\% of UQ). A substantial aspect nonetheless disregarded was the language-specific statistics, especially those regarding \textit{question patterns}. As shown in Table \ref{stat}, for both MCWQ and MCWQ-R, we observe a decrease in question patterns in translations compared with English and the corresponding pairs coupled with SPARQL queries, i.e., question-query pairs. This indicates that the patterns are partially collapsed in the target languages with both methodologies. Furthermore, as the SPARQL queries are invariant logical representations underlying the semantics, the QA pairs are supposed to be consistent with the question patterns even if collapsed. However, we notice a significant inconsistency ($\Delta_{JA}=240$; $\Delta_{ZH}=578$) between the two items in MCWQ while there are few differences ($\Delta_{JA}=0$; $\Delta_{ZH}=9$) in MCWQ-R. This further implicates a resultant disconnection between the translated questions and corresponding semantic representations with NMT.

We expect our grammar to be fully deterministic over the dataset, nonetheless, it fails to disambiguate a small proportion (322; 0.31\%) of English utterance patterns that are \textit{amphibologies} (grammatically ambiguous) and requires reasoning beyond the scope of grammar. We let the model randomly assign a candidate translation for these.

\subsection{Translation Quality Assessment}\label{quality}
Following \citet{cui-etal-2022-compositional}, we comprehensively assess the translation quality of MCWQ-R and the GT counterpart based on the \textit{test-intersection} set (the intersection of the test sets of all splits) samples. While translation quality is a general concept, in this case, we focus on how appropriately the translation trades off fluency and faithfulness to the principle of compositionality.

\begin{table}[htbp]
\centering
\resizebox{\columnwidth}{!}{
\small
\begin{tabular}{@{}cccccc@{}}
\toprule
        \multicolumn{2}{c}{Language}        & Reference & \multicolumn{3}{c}{Manual} \\ \cmidrule(r){3-3} \cmidrule(r){4-6}
\multicolumn{2}{c}{\& Method}          & BLEU      & avgMP         & avgF     & $\mathrm{P(MP, F\ge{3})}$      \\ \midrule
\multirow{2}{*}{JA} & RBMT  & 97.1  & 4.8   & 4.0  & 100.0\%  \\ 
&GT & 45.1      & 3.7      & 4.1   & 71.4\%     \\
\midrule
\multirow{2}{*}{ZH} & RBMT  & 94.4  & 4.9   & 4.2  & 100.0\%  \\ 
&GT & 47.2      & 3.6      & 4.2   & 71.4\%     \\
\bottomrule
\end{tabular}
}
\caption{\label{tbl:assess}
Assessment scores for the translations. \textbf{MP} refers to Meaning Preservation and \textbf{F} refers to Fluency. The prefix \textbf{avg} indicates averaged scores. $\mathrm{P(MP, F\ge{3})}$ refers to the proportion of questions regarded as \textit{acceptable}.}
\end{table}

\paragraph{Reference-based assessment.}
We manually translate 155 samples from the \textit{test-intersection} set in a faithful yet \textit{rigid} manner as gold standard before the grammar construction. We calculate BLEU \citep{BLEU} scores of the machine-translated questions against the gold set with sacreBLEU \citep{sacreBLEU}, shown in Table~\ref{tbl:assess}. Our RBMT reached 97.1 BLEU for Japanese and 94.4 for Chinese, indicating a nearly perfect translation as expected. While RBMT could ideally reach a full score, the loss here is mainly caused by samples lacking context information (agnostic of entity for instance). In addition, we observe that GT obtained fairly poor performance with 45.1 BLEU for Japanese, which is significantly lower than the other branches in MCWQ (87.4, 76.6, and 82.8 for Hebrew, Kannada, and Chinese, respectively; \citealp{cui-etal-2022-compositional}). The main reason for this gap is the different manner in which we translated the gold standard: the human translators in MCWQ took a looser approach.

\paragraph{Manual assessment.}
We manually assess the translations of 42 samples (for each structural complexity level defined by \citealp{CFQ}) in terms of \textit{meaning preservation} (MP) and \textit{fluency} (F) with a rating scale of 1--5. As shown in Table \ref{tbl:assess}, our translations have significantly better MP than GT, which is exhibited by the average scores (1.1 and 1.3 higher in avgMP for Japanese and Chinese, respectively). However, the methods obtain similar fluency scores, indicating that both suffer from unnatural translations, partially because of the unnaturalness of original English questions \citep{cui-etal-2022-compositional}. RBMT produces only few translations with significant grammar errors and semantic distortions, while GT results in 28.6\% of unacceptable translations in this respect. Such errors occur on similar samples for the two languages, suggesting a systematicity in GT failure. We include details of manual assessment in Appendix \ref{appendix:assess}.

\section{Experiments}\label{experiments}
While extensive experiments have been conducted on both the monolingual English \cite{CFQ} and the GT-based multilingual benchmarks \cite{cui-etal-2022-compositional}, the results fail to demonstrate pure multilingual CG due to noisy translations. Consistent with prior work, we experiment in both monolingual and cross-lingual scenarios. Specifically, we take into consideration both RBMT and GT branches\footnote{The GT-Chinese data (and part of the corresponding results) is from MCWQ (released under the CC-BY license). The GT-Japanese is generated following the same pipeline.} in the experiments for further comparison. 

\subsection{Within-language Generalization (Monolingual)}
\citet{cui-etal-2022-compositional} showed consistent ranking among sequence-to-sequence (seq2seq) models for the 4 splits (3 MCD and 1 random splits). We fine-tune and evaluate the pre-trained mT5-small \cite{xue-etal-2021-mt5}, which performs well on MCWQ for each monolingual dataset. In addition, we train a model using mBART50 \citep{tang2020multilingual} as a frozen embedder and learned Transformer encoder and decoder, following \citet{liu-etal-2020-multilingual-denoising}. We refer to this model as mBART50$^{*}$ (it is also the base architecture of ZX-Parse; see~\S\ref{sec:xl}).

We show the monolingual experiment results in Table \ref{monolingual}.
The models achieve better average performance on RBMT questions than GT ones. This meets our expectations since the systematically translated questions excluded the noise. On the random split, both RBMT branches are highly consistent with English, while noise in GT data lowers accuracy. However, the comparisons on MCD splits show that RBMT branches are less challenging than English, especially for mT5-small. In~\S\ref{mono-analysis}, we show this is due to the ``simplifying'' effect of translation on composition.

Comparisons across languages demonstrate another interesting phenomenon: Japanese and Chinese exhibited an \textit{opposite} relative difficulty on RBMT and GT. It is potentially due to the more extensive grammatical system (widely applied in different realistic scenes) of the Japanese language, while the grammatical systems and language styles are unified in RBMT, the GT tends to infer such diversity which nonetheless belongs to another category \citep[natural language variant;][]{shaw-etal-2021-compositional}.

\begin{table}[htb]
\centering
\resizebox{0.48\textwidth}{!}{
\begin{tabular}{cc||cc|cc} 
\toprule
\multicolumn{2}{l}{Exact} & \multicolumn{2}{c}{\texttt{mT5-small}}                         & \multicolumn{2}{c}{\texttt{mBART50$^{*}$}}                      \\ 
\cmidrule(r){3-4}  \cmidrule(r){5-6} 
\multicolumn{2}{l}{Match(\%)}   & \multicolumn{1}{l}{MCWQ-R} & \multicolumn{1}{l}{MCWQ} & \multicolumn{1}{l}{MCWQ-R} & \multicolumn{1}{l}{MCWQ}  \\ 
\midrule
\multirow{3}{*}{\transbox{MCD$_{\textrm{mean}}$}} & EN                               & \multicolumn{2}{c|}{38.3}                              & \multicolumn{2}{c}{$ 55.2_{\pm1.6}$}                       \\ 
                                       & JA                               & 56.3                       & 30.8                     & 58.3                       & 32.9                      \\ 
                                       & ZH                               & 51.1                       & 36.3                     & 59.9                       & 43.6                      \\ 
\midrule \midrule
\multirow{3}{*}{\transbox{Random}}                & EN                               & \multicolumn{2}{c|}{98.6}                              & \multicolumn{2}{c}{$ 98.9_{\pm0.1}$}                       \\ 
                                       & JA                               & 98.7                       & 92.4                     & 98.7                       & 92.9                      \\ 
                                       & ZH                               & 98.4                       & 91.8                     & 98.8                       & 92.8                      \\
\bottomrule
\end{tabular}
}
\caption{\label{monolingual}Monolingual experiment results: Exact match accuracies in percentage (\%) are shown here. We present the model performance on the two translated datasets, which share the English branch. MCD$_{\textrm{mean}}$ represents the average accuracy across 3 MCD splits, and the detailed results breakdown can be found in Appendix \ref{appendix: mcds}. Random refers to the results on the random split. We run 3 replicates for mBART50$^{*}$ on EN, which is used for further cross-lingual experiments (see \S\ref{sec:xl}). 
}
\end{table}

\subsection{Cross-lingual Generalization (Zero-shot)}\label{sec:xl}
We mentioned the necessity of developing multilingual KBQA systems in \S\ref{intro}. Enormous efforts required for model training for every language encourage us to investigate the zero-shot cross-lingual generalization ability of semantic parsers which serve as the KBQA backbone. While similar experiments were conducted by \citet{cui-etal-2022-compositional}, the adopted pipeline (cross-lingual inference by mT5 fine-tuned on English) exhibited negligible predictive ability for all the results, from which we can hardly draw meaningful conclusions.

For our experiments, we retain this as a baseline, and additionally train Zero-shot Cross-lingual Semantic Parser (ZX-Parse), a multi-task seq2seq architecture proposed by \citet{sherborne-lapata-2022-zero}. The architecture consists of mBART50$^{*}$ with two auxiliary objectives (question reconstruction and language prediction) and leverages \textit{gradient reversal} \cite{ganin2016domain} to align multilingual representations, which results in a promising improvement in cross-lingual SP.

With the proposed architecture, we investigate how the designed cross-lingual parser and its representation alignment component perform on the compositional data. Specifically, we experiment with both the full ZX-Parse and with mBART50$^{*}$, its logical-form-only version (without auxiliary objectives). For the auxiliary objectives, we use bi-text from MKQA \citep{longpre-etal-2021-mkqa} as supportive data. See Appendix \ref{appendix:train} for details.

Table \ref{crosslingual} shows our experimental results. mT5-small fine-tuned on English fails to generate correct SPARQL queries. ZX-Parse, with a frozen mBART50 encoder and learned decoder, demonstrates moderate predictive ability. Surprisingly, while the logical-form-only (mBART50$^{*}$) architecture achieves fairly good performance both within English and cross-lingually, the auxiliary objectives cause a dramatic decrease in performance. We discuss this in \S\ref{XL-analysis}

\begin{table*}[th]
\centering
\resizebox{0.75\textwidth}{!}{

\begin{tabular}{cc||cc|cc|cc} 
\toprule
\multicolumn{2}{l}{Exact} & \multicolumn{2}{c}{\texttt{mT5-small}}                         & \multicolumn{2}{c}{\texttt{mBART50$^*$}}       & \multicolumn{2}{c}{\texttt{ZX-Parse}}                           \\ \cmidrule(r){3-4}  \cmidrule(r){5-6}  \cmidrule(r){7-8}
\multicolumn{2}{l}{Match(\%)}  & \multicolumn{1}{l}{MCWQ-R} & \multicolumn{1}{l}{MCWQ} & \multicolumn{1}{l}{MCWQ-R} & \multicolumn{1}{l}{MCWQ} & \multicolumn{1}{l}{MCWQ-R} & \multicolumn{1}{l}{MCWQ}  \\ 
\midrule
\multirow{3}{*}{MCD$_{\textrm{mean}}$} & \textit{\textcolor{gray}{EN}}                                          & \multicolumn{2}{c|}{\textcolor{gray}{38.3}}                              & \multicolumn{2}{c|}{\textcolor{gray}{$ 55.2_{\pm1.6}$}}                    & \multicolumn{2}{c}{\textcolor{gray}{$ 23.9_{\pm3.4}$}}                     \\ 
                                       & JA                                                   & 0.10                       & 0.14                     & $ 35.4_{\pm2.1}$             & $ 24.6_{\pm2.8}$           & $ 8.8_{\pm1.8}$              & $ 8.5_{\pm1.5}$             \\ 
                                       & ZH                                                   & 0.12                       & 0.18                     & $ 37.7_{\pm1.8}$             & $ 35.0_{\pm2.2}$           & $ 9.3_{\pm2.0}$              & $ 9.1_{\pm1.7}$             \\ 
\midrule
\multirow{3}{*}{Random}                & \textit{\textcolor{gray}{EN}}                                          & \multicolumn{2}{c|}{\textcolor{gray}{98.6}}                              & \multicolumn{2}{c|}{\textcolor{gray}{$ 98.9_{\pm0.1}$}}                    & \multicolumn{2}{c}{\textcolor{gray}{$ 75.9_{\pm9.1}$}}                     \\ 
                                       & JA                                                   & 0.9                        & 0.9                      & $ 58.0_{\pm0.8}$             & $ 34.4_{\pm3.1}$           & $ 27.2_{\pm2.1}$             & $ 23.1_{\pm1.9}$            \\ 
                                       & ZH                                                   & 1.4                        & 1.1                      & $ 58.2_{\pm1.4}$             & $ 43.7_{\pm1.3}$           & $ 29.4_{\pm3.4}$             & $ 24.8_{\pm3.5}$            \\
\bottomrule
\end{tabular}}
\caption{\label{crosslingual}
Cross-lingual experiment results. The \textit{English} results in \textcolor{gray}{gray} refer to within-language generalization performance. Notice that mBART50$^*$ here is the ablation model of ZX-Parse with the same training paradigm for logical form decoder. We run 3 replicates for mBART50$^{*}$ and ZX-Parse. The results breakdown for 3 MCD splits can be found in Appendix \ref{appendix: mcds}.
}
\end{table*}

\section{Discussion}\label{analysis}
\subsection{Monolingual Performance Gap}\label{mono-analysis}
As Table \ref{monolingual} suggests, MCWQ-R is easier than its English and GT counterparts. While we provide evidence that the latter suffers from translation noise, comparison with the former indicates partially degenerate compositionality in our multilingual sets. We ascribe this degeneration to an inherent property of translation, resulting from linguistic differences: as shown in Table \ref{stat}, question patterns are partially collapsed after mapping to target languages. 

\paragraph{Train-test overlap.}
Intuitively, we consider training and test sets of the MCD splits, where no overlap is permitted in English under MCD constraints (the train-test intersection must be empty). Nevertheless, we found such overlaps in Japanese and Chinese due to the collapsed patterns. Summing up over 3 MCD splits, we observe 58 samples for Japanese and 37 for Chinese, and the two groups share similar patterns.
Chinese and Japanese grammar inherently fail to (naturally) express specific compositions in English, predominantly the \textit{possessive case}, a main category of compositional building block designed by \citet{CFQ}. This linguistic divergence results in degeneration in compound divergence between training and test sets, which is intuitively reflected by the pattern overlap. We provide examples in Appendix \ref{appendix:simplification}.


\paragraph{Loss of structural variation.}
Given the demonstration above, we further look at MCWQ and see whether GT could avoid this degeneration. Surprisingly, the GT branches have larger train-test overlaps (108 patterns for Japanese and 144 for Chinese) than RBMT counterparts , among which several samples (45 for Japanese and 55 for Chinese) exhibit the same structural collapse as in RBMT. Importantly, a remaining large proportion of the samples (63 for Japanese and 89 for Chinese) possess different SPARQL representations for training and test respectively. In addition, several ill-formed samples are observed in this intersection. 

The observations above provide evidence that the structural collapse is due to \textit{inherent} linguistic differences and thus generally exists in translation-based methods, resulting in compositional degeneration in multilingual benchmarks. For GT branches, the noise involving semantic and grammatical distortion dominates over the degeneration, and thus causes worse model performance.

\paragraph{Implications.}
While linguistic differences account for the performance gaps, we argue that monolingual performance in CG cannot be fairly compared across languages with translated benchmarks. While ``translationese'' occurs in translated datasets for other tasks too \cite{riley-etal-2020-translationese,bizzoni-lapshinova-koltunski-2021-measuring,vanmassenhove-etal-2021-machine}, it is particularly significant here.

\subsection{Cross-lingual Generalization}\label{XL-analysis}


 \begin{figure}[tb]
	\centering
	\includegraphics[width=70mm]{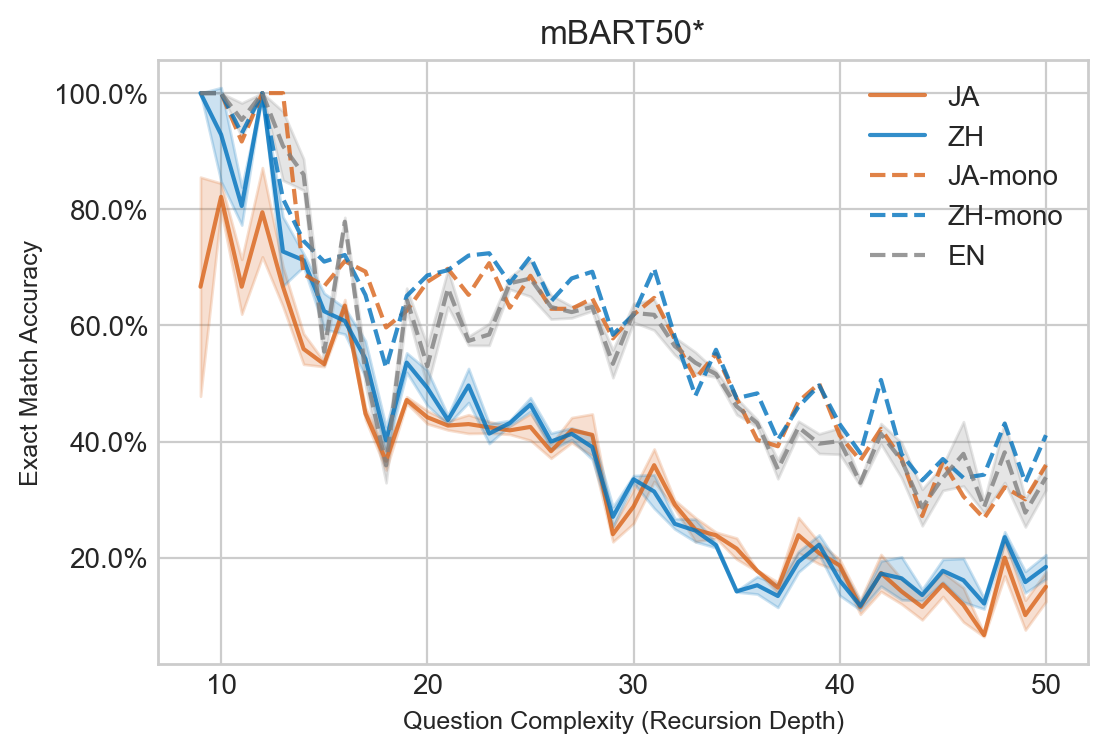}
\caption{\label{acc.vs.cplx}
Accuracy on \textbf{MCWQ-R} of mBART50$^{*}$ varies against increasing question complexity, averaged over 3 MCD splits. Dashed lines refer to within-language generalization performance, indicating cross-lingual transfer upper boundaries.
}	
\end{figure}

\paragraph{PLM comparison.}
mT5 fine-tuned on English fails to generalize cross-lingually (Table \ref{crosslingual}). ZX-Parse, based on mBART50, achieved fair performance. A potential reason is that mT5 (especially small and base models) tends to make ``accidental translation'' errors in zero-shot generalization \cite{xue-etal-2021-mt5}, while the representation learned by mBART enables effective unsupervised translation via language transfer \cite{liu-etal-2020-multilingual-denoising}. 
Another surprising observation is that mBART50$^{*}$ outperforms the fine-tuned mT5-small on monolingual English (55.2\% for MCD$_{\textrm{mean}}$) with less training. We present additional results regarding PLM fine-tuning in Appendix \ref{appendix:additional}.

\paragraph{Hallucination in parsing.} 
mT5 tends to output partially correct SPARQL queries due to its drawback in zero-shot generative scenarios. From manual inspection, we note a common pattern in these errors that can be categorized as \textit{hallucinations} \cite{survey_halluc, guerreiro-etal-2023-looking}. As Table \ref{tab:halluc.percent} suggests, the hallucinations with \texttt{country} entities occur in most wrong predictions, and exhibit a \textit{language bias} akin to that \citet{kassner-etal-2021-multilingual} found in mBERT \cite{devlin-etal-2019-bert}, i.e., mT5 tends to predict the country of origin associated with the input language in the hallucinations, as demonstrated in Table \ref{example_halluc}. Experiments in Appendix \ref{appendix:additional} indicate that the bias is potentially encoded in the pre-trained decoders.

\begin{table}[htb]
    \centering
    {\small
    \begin{tabular}{ll|ccc|ccc}
    \toprule
    \multicolumn{2}{l}{\textbf{Halluc.(\%)}} & \multicolumn{3}{c}{MCD$_{\textrm{mean}}$} & \multicolumn{3}{c}{Random} \\
                            \cmidrule(r){3-5} \cmidrule(r){6-8}
     \multicolumn{2}{l}{W/ \texttt{country}} & ZH & JA & \textit{\textcolor{gray}{EN}} & ZH & JA & \textit{\textcolor{gray}{EN}} \\
    \midrule
     \texttt{Q148} & \texttt{CN} & \textbf{71.0} & 0 & \textcolor{gray}{0} & \textbf{60.6} & 0 & \textcolor{gray}{0} \\
     \texttt{Q17} & \texttt{JP} & 0.1 & \textbf{76.1} & \textcolor{gray}{0} & 0.1 & \textbf{63.3} & \textcolor{gray}{0} \\
     \multicolumn{2}{c|}{\textit{Others}} & 4.2 & 1.8 & \textcolor{gray}{0.45} & 3.8 & 0.9 & \textcolor{gray}{0} \\
     \midrule
     \multicolumn{2}{c|}{Total} & 75.2 & 77.9 & \textcolor{gray}{0.45} & 64.4 & 64.2 & \textcolor{gray}{0} \\
     \bottomrule
    \end{tabular}
    }
    \caption{Proportion of hallucinations with the specific \texttt{country} entities in the wrong predictions, generated by mT5-small in zero-shot cross-lingual generalization (models trained on English). Within-language results are in \textcolor{gray}{gray} for comparison. The results on MCWQ-R are shown here. The countries are represented in QID and ISO codes, and the other (12) countries involved in the dataset are summed as \textit{others}. The predominant parts exhibiting language bias are in \textbf{bold}, for which an example is shown in Table \ref{example_halluc}.}
    \label{tab:halluc.percent}
\end{table}

\begin{table}[ht]
{\footnotesize \centering
\begin{tabular}{p{0.14\textwidth}p{0.28\textwidth}}
\toprule
\textbf{Question (EN)} & Which actor was M0 's actor \\
\midrule

\textbf{Question (ZH)}  & \begin{CJK*}{UTF8}{gbsn} M0的演员是哪个演员 \end{CJK*}\\
\textbf{Inferred (RIR)} & {\tt \footnotesize
\textbf{SELECT DISTINCT} ?x0 \textbf{WHERE} lb ( M0 ( wdt:P453 ) ( ?x0 ) ) . \textcolor{red}{( ?x0 ( wdt:P27 ) ( wd:Q148 ) )} rb} \\

\midrule

\textbf{Question (JA)} & \begin{CJK*}{UTF8}{min} M0の俳優はどの俳優でしたか \end{CJK*}\\ 
\textbf{Inferred (RIR)} & {\tt \footnotesize
\textbf{SELECT DISTINCT} ?x0 \textbf{WHERE} lb ( ?x0 ( wdt:P106 ) ( wd:Q33999 ) ) . ( M0 ( wdt:P108 ) ( ?x0 ) ) . \textcolor{red}{( ?x0 ( wdt:P27 ) ( wd:Q17 ) )} rb}\\
\bottomrule
\end{tabular}}
\caption{\label{example_halluc}
An example of the language-biased hallucinations. The questions are parallel across languages and associated with the same SPARQL query. The inferred queries are in RIR form. The language-biased hallucination triples are highlighted in red, where \texttt{\textcolor{red}{Q148}} is \texttt{China} in Wikidata, and \texttt{\textcolor{red}{Q17}} is \texttt{Japan}.
}
\end{table}

\paragraph{Representation alignment.}
The auxiliary objectives in ZX-Parse are shown to improve the SP performance on MultiATIS++ \citep{xu-etal-2020-end} and Overnight \citep{wang-etal-2015-building}. However, it leads to dramatic performance decreases on all MCWQ and MCWQ-R splits. We include analysis in Appendix \ref{appendix:align}, demonstrating the moderate effect of the alignment mechanism here, which nevertheless should reduce the cross-lingual transfer penalty. We thus ascribe this gap to the natural utterances from MKQA used for alignment resulting in less effective representations for compositional utterances, and hence the architecture fails to bring further improvement.

 
\paragraph{Cross-lingual difficulty.}
As illustrated in Figure \ref{acc.vs.cplx}, while accuracies show similar declining trends across languages, cross-lingual accuracies are generally closer to monolingual ones in low complexity levels, which indicates that the cross-lingual transfer is difficult in CG largely due to the failure in universally representing utterances of high compositionality across languages. 
Specifically, for low complexity samples, we observe test samples that are correctly predicted cross-lingually but wrongly predicted within English. These several samples (376 for Japanese and 395 for Chinese on MCWQ-R) again entail structural simplification, which further demonstrates that this eases the compositional challenge even in the cross-lingual scenario.
We further analyze the accuracies by complexity of MCWQ and ZX-Parse in Appendix \ref{appendix:acc.by.complexity}.

\section{Conclusion}\label{conclusion}
In this paper, we introduced MCWQ-R, a robustly generated multilingual CG benchmark with a proposed rule-based framework.
Through experiments with multilingual data generated with different translation methods, we revealed the substantial impact of linguistic differences and ``translationese'' on compositionality across languages. Nevertheless, removing of all difficulties but compositionality, the new benchmark remains challenging both monolingually and cross-lingually.
Furthermore, we hope our proposed method can facilitate future investigation on multilingual CG benchmark in a controllable manner.

\section*{Limitations}
Even the premise of parsing questions to Wikidata queries leads to linguistic and cultural bias, as Wikidata is biased towards English-speaking cultures \citep{10.1145/3484828}. As \citet{cui-etal-2022-compositional} argue, speakers of other languages may care about entities and relations that are not represented in English-centric data \cite{liu-etal-2021-visually,hershcovich-etal-2022-challenges}. For this reason and for the linguistic reasons we demonstrated in this paper, creating CG benchmarks natively in typologically diverse languages is essential for multilingual information access and its evaluation.

As we mentioned in \S\ref{stats}, our translation system fails to deal with ambiguities beyond grammar and thus generates wrong translations for a few samples (less than 0.31\%). Moreover, although the dataset can be potentially augmented with low-resource languages and in general other languages through the translation framework, adequate knowledge will be required to expand rules for the specific target languages.

With limited computational resources, we are not able to further investigate the impact of parameters and model sizes of multilingual PLM as our preliminary results show significant performance gaps between PLMs.

\section*{Broader Impact}
A general concern regarding language resource and data collection is the potential (cultural) bias that may occur when annotators lack representativeness. Our released data largely avoid such issue due to the synthetic and cultural-invariant questions based on knowledge base. Assessment by native speakers ensures its grammatical correction. However, we are aware that bias may still exist occasionally. For this purpose, we release the toolkit and grammar used for generation, which allows further investigation and potentially generating branches for other languages, especially low-resource ones.

In response to the appeal for greater environmental awareness as highlighted by \citet{hershcovich-etal-2022-towards}, a climate performance model card for mT5-small is reported in Table \ref{tbl:climate_model_card}. By providing access to the pre-trained models, we aim to support future endeavors while minimizing the need for redundant training efforts.
\begin{table}[htb]
\small
\adjustbox{width=\columnwidth}{
\begin{tabular}{@{}p{55mm}p{25mm}@{}}
\toprule
\multicolumn{2}{c}{\textbf{\texttt{mT5-small} finetuned}} \\
\midrule
1. Model publicly available? & Yes\\
2. Time to train final model & 21 hours\\
3. Time for all experiments & 23 hours\\
4. Energy consumption & 0.28kW\\
5. Location for computations & Denmark\\
6. Energy mix at location & 191gCO2eq/kWh\\
7. CO2eq for final model & 4.48 kg\\
8. CO2eq for all experiments & 4.92 kg\\
\bottomrule
\end{tabular}
}
\caption{Climate performance model card for mT5-small fine-tuned on MCWQ/MCWQ-R. ``Time to train final model'' corresponds to the training time for a single model of one split and one language, while the remaining models have similar resource consumption.}
\label{tbl:climate_model_card}
\end{table}

\section*{Acknowledgements}
We thank the anonymous reviewers for their valuable feedback. We are also grateful to Guang Li, Nao Nakagawa, Stephanie Brandl, Ruixiang Cui, Tom Sherborne and members of the CoAStaL NLP group for their helpful insights, advice and support throughout this work.

\bibliography{anthology,custom}
\bibliographystyle{acl_natbib}

\appendix

\section{Transduction Grammar Examples}\label{appendix:grammar}
\paragraph{Inflection in Japanese.}
We provide a concrete example regarding the linguistic divergences during translation and how our transduction grammar (SCFG) address it. We take Japanese, specifically its verbal \textit{inflection} case as an example.
\\ \hspace*{\fill} \\
\begin{CJK}{UTF8}{min}
\noindent \textbf{GRAMMAR}
\begin{equation}
\label{grammar}
\begin{aligned}
\mathrm{VP} \ \rightarrow \ &\langle \mathrm{V \ NP},  \ \mathrm{NP \ V \rangle}\\
\mathrm{V} \ \rightarrow \ &\langle \mathrm{VT \ andV},  \ \mathrm{VT \ andV \rangle}\\
\mathrm{andV} \ \rightarrow \ &\langle \mathrm{and \ V},  \ \mathrm{\epsilon \ V \rangle}\\
\mathrm{NP} \ \rightarrow \ &\langle \text{a film},  \  \text{映画} \rangle\\
\mathrm{V} \ \rightarrow \ &\{\langle \text{edit \ , 編集します} \rangle ,\\ &\langle \text{write \ , 書きます} \rangle \} \\
\mathrm{VT} \ \rightarrow \ &\{\langle \text{edit \ , 編集し} \rangle ,\\ &\langle \text{write \ , 書き} \rangle \}\\
\end{aligned}
\end{equation}
\textbf{GENERATED STRING}
\begin{equation}
\label{sync-str}
\begin{aligned}
\langle \text{\textcolor{orange}{write} and \textcolor{blue}{edit} a film}, \ &\text{映画を \textcolor{orange}{書き} \textcolor{blue}{編集します}} \rangle \\ 
\langle \text{\textcolor{blue}{edit} and \textcolor{orange}{write} a film}, \ &\text{映画を \textcolor{blue}{編集し} \textcolor{orange}{書きます}} \rangle
\end{aligned}
\end{equation}
\end{CJK}

\begin{CJK}{UTF8}{min}
In the string pair of (\ref{sync-str}), the Japanese verbal inflection is reasoned from its position in a sequence where correspondences are highlighted with different colors. To make it more intuitive, consider a phrase (out of the corpus) ``\textit{run and run}'' with repeated verb ``run'' and its Japanese translation "\textcolor{olive}{\ruby{走}{hashi}\ruby{り}{ri}}、\textcolor{olive}{\ruby{走}{hashi}\ruby{り}{ri}}\textcolor{purple}{\ruby{ま}{ma}\ruby{す}{su}}", where the repeated "\textcolor{olive}{\ruby{走}{hashi}\ruby{り}{ri}}"(which should belong to V if in (\ref{grammar})) refers to a category of verb base, namely \textit{conjunctive} indicating that it could be potentially followed by other verbs\footnote{Formally, the conjunctive in Japanese involves 2 forms: chushi-form and te-form, to keep consistent with the English questions (where temporal ordering is not entailed by coordination), we adopt the former form in our grammar since it indicates weaker temporal ordering than the latter \cite{conjunctivejp}.}; and the inflectional suffix "\textcolor{purple}{\ruby{ま}{ma}\ruby{す}{su}}" indicting the end of the sentence. Briefly speaking, in the Japanese grammar, the last verb in a sequence have a different form from the previous ones depending on the formality level.

In this case, the transduction rule of the lowest syntactic level explaining this inflection is $\mathrm{V} \ \rightarrow \ \langle \mathrm{VT \ andV},  \ \mathrm{VT \ andV \rangle}$, therefore the \text{VT} with \textit{suffix} T is derived from \text{V} (V exhibit no inflection regarding ordering in English) from this level and carries this context information down to the terminals. Considering questions with deep parse trees where such context information should potentially be carried through multiple part-of-speech symbols in the top-down process, we let the \textit{suffix} be \textit{inheritable} as demonstrated in (\ref{inflection}). 
\begin{equation}
\label{inflection}
\begin{aligned}
\mathrm{VP} \ \rightarrow \ &\langle \mathrm{VP{\color{blue}T} \ andVP},  \ \mathrm{VP{\color{blue}T} \ andVP \rangle}\\
\mathrm{VP{\color{blue}T}} \ \rightarrow \ &\langle \mathrm{V{\color{blue}T} \ NP},  \ \mathrm{NP \ V{\color{blue}T} \rangle}
\end{aligned}
\end{equation}
where suffix T carries the commitment of inflection to be performed at the non-terminal level and is explained by context of VPT and inherited by VT. While such suffix is commonly used in formal grammar, we leverage this mechanism to a large extent to fill the linguistic gap. The strategy is proved to be simple yet effective in practical grammar construction to handle most of the problems caused by linguistic differences such as inflection as mentioned.

\end{CJK}

\section{Translation Assessment Details}\label{appendix:assess}
Since manual assessment is subjective, the guidelines were stated before assessment: translations resulting in changed expected answer domains are rated 1 or 2 for \textit{meaning preservation}. Those with major grammar errors are rated 1 or 2 for \textit{fluency}. Accordingly, we regard questions with a score $\geq{3}$ as acceptable in the corresponding aspect. 

\begin{figure}[tb]
    \centering
    \includegraphics[width=0.5\textwidth]{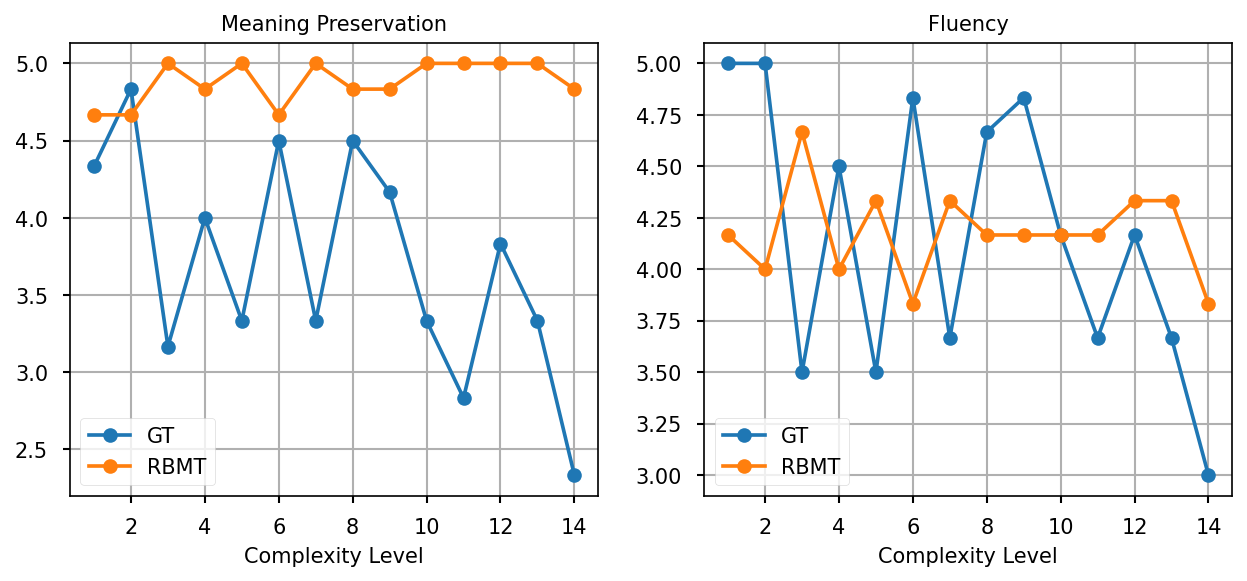}
    \caption{Manual assessment scores vary against increasing complexity levels with a bin size of 3. The scores are averaged over every 3 complexity levels and 2 languages.}
    \label{fig:manassess}
\end{figure}

To make an intuitive comparison, we divide the 42 complexity levels (for each level we sampled 1 sentence) into 14 coarser levels and see the variation of the scores of 2 methods against the increasing complexity. As shown in Figure \ref{fig:manassess}, Our method exhibits uniformly good meaning preservation ability while GT suffers from semantic distortion for certain cases and especially for those of high complexity. For the variation of fluency, the steady performance of our method indicates that the loss is primarily \textit{systematic} and due to compromise for compositional consistency and parallel principle, while GT generates uncontrollable results with incorrect grammar (and thus illogical) occasionally.
\begin{CJK}{UTF8}{min}
We present imprecise translation example of our method. Adjective indicating nationalities such as \textit{``American''} is naturally adapted to \textit{``\ruby{ア}{a}\ruby{メ}{me}\ruby{リ}{ri}\ruby{カ}{ka}\ruby{人}{jin}(American person)''} when modifying a person in Japanese; then for a sample (note that entities are bracketed):
\\[3pt]
\noindent \textbf{Input}:\textit{``Was [Kate Bush] \textcolor{orange}{British}''}

\noindent \textbf{Output}:\textit{``[Kate Bush]\ruby{は}{wa}\textcolor{orange}{\ruby{イ}{i}\ruby{ギ}{gi}\ruby{リ}{ri}\ruby{ス}{su}}\ruby{の}{no}\ruby{で}{de}\ruby{し}{shi}\ruby{た}{ta}\ruby{か}{ka}''}

\noindent \textbf{Expected}:\textit{``[Kate Bush]\ruby{は}{wa}\textcolor{orange}{\ruby{イ}{i}\ruby{ギ}{gi}\ruby{リ}{ri}\ruby{ス}{su}}\textcolor{blue}{\ruby{人}{jin}}\ruby{で}{de}\ruby{し}{shi}\ruby{た}{ta}\ruby{か}{ka}''}
\\[3pt]
Consider the bracketed entity [Kate Bush] which is invisible during translation, and also the fact that the sentence still holds if it is alternated with non-human entities. Without the contribution of the entity semantics, the grammar is unable to specify \textit{``\textcolor{blue}{\ruby{人}{jin}}(person)''} in this case, and results in a less natural expression. We observed a few samples similar to this leading to the error in BLEU scores.

For GT, as we mentioned in \S\ref{quality}, it causes semantic distortions potentially changing expected answers:
\\[3pt]
\noindent \textbf{Input}:\textit{``\textit{What did [human] \textcolor{orange}{found}}''}

\noindent \textbf{Output (GT)}:\textit{``[human] \ruby{は}{wa}\ruby{何}{nani}\ruby{を}{wo}\textcolor{blue}{\ruby{見}{mi}\ruby{つ}{tsu}\ruby{け}{ke}\ruby{ま}{ma}\ruby{し}{shi}\ruby{た}{ta}}\ruby{か}{ka}''}

\noindent \textbf{Expected (\&Ours)}:\textit{``\textit{[human]\ruby{が}{ga}\textcolor{orange}{\ruby{創}{so}\ruby{設}{setsu}\ruby{し}{shi}\ruby{た}{ta}}\ruby{の}{no}\ruby{は}{wa}\ruby{何}{nan}\ruby{で}{de}\ruby{す}{su}\ruby{か}{ka}''}}
\\[3pt]
Disregarding the sentence patterns, the output of GT distorted the meaning as \textit{``What did [human] \textcolor{blue}{find}''}, translated back to English. 
\\[3pt]
\noindent \textbf{Input}:\textit{``\textit{Was a \textcolor{orange}{prequel} of [Batman: Arkham Knight] 's \textcolor{orange}{prequel}...}''}

\noindent \textbf{Output (GT)}:\textit{``[Batman: Arkham Knight] \ruby{の}{no}\textcolor{orange}{\ruby{前}{zen}\ruby{日}{jitsu}\ruby{譚}{tan}}...''}

\noindent \textbf{Expected (\&Ours)}:\textit{``\textit{[Batman: Arkham Knight] \ruby{の}{no}\textcolor{orange}{\ruby{前}{zen}\ruby{日}{jitsu}\ruby{譚}{tan}}\ruby{の}{no}\textcolor{orange}{\ruby{前}{zen}\ruby{日}{jitsu}\ruby{譚}{tan}}...}''}
\\[3pt]
The example above shows how the 2 methods deal with a compositional phrase occurring in the dataset. GT exhibits reasoning ability which understood that \textit{``a prequel of a prequel''} indicates \textit{``a prequel''} thus translating it as \textit{``\ruby{前}{zen}\ruby{日}{jitsu}\ruby{譚}{tan}(prequel)''}, whereas an expected compositionally faithful translation should be \textit{``\ruby{前}{zen}\ruby{日}{jitsu}\ruby{譚}{tan}\ruby{の}{no}\ruby{前}{zen}\ruby{日}{jitsu}\ruby{譚}{tan}(a prequel of a prequel)''}. The examples demonstrate how GT as a neural model fails in accommodating compositionality even for the well-formed translations: the \textit{infinite} compositional expression potentially reaches the ``fringe area'' of the trained neural model distribution, i.e., it overly concerns the possibility that the sentence occurs instead of keeping faithful regarding the atoms and their compositions.
\end{CJK}

\section{Training Details}\label{appendix:train}
\paragraph{mT5-small.} 
We follow the same setup of mT5-small as in \cite{cui-etal-2022-compositional} with default hyperparameters but a learning rate of $5e^{-4}$, which is believed to help overcome the local minimum. Each model was trained on 4 Titan RTX GPUs with a batch size of 16. The total training time is 234 hours for 12 models (4 splits for GT-Japanese, RBMT-Chinese and RBMT-Japanese respectively).

\paragraph{mBART50 and ZX-Parse.}
We follow the searched optimal architecture and parameters\footnote{Specifically the configuration provided in \url{https://github.com/tomsherborne/zx-parse}} by \citet{sherborne-lapata-2022-zero}. The logical-form-only mBART50$^{*}$ comprises frozen mBART50-large embedder, 1-layer encoder, and 6-layer decoder, and the full ZX-Parse with additional alignment components: 6-layer decoder (reconstruction) and 2-layer feed-forward networks (language prediction) trained with bi-text that we extract from MKQA. The auxiliary components in ZX-Parse make the encoder align latent representations across languages. Each model was trained on 1 Titan RTX GPU with a batch size of 2. It takes around 17 hours to train a full ZX-Parse and 14 hours an mBART50$^{*}$ model.

\section{Additional Results}
\subsection{MCD Splits}\label{appendix: mcds}
The exact match accuracies on the 3 maximum compound divergence (MCD) splits \cite{CFQ} are shown in Table \ref{tab:mcds}.

\begin{table*}[tb]
    \centering
    \resizebox{0.75\textwidth}{!}{
\begin{tabular}{cc||cc|cc|cc} 
\toprule
\multicolumn{2}{l}{Exact} & \multicolumn{2}{c}{\texttt{mT5-small}}                         & \multicolumn{2}{c}{\texttt{mBART50$^*$}}       & \multicolumn{2}{c}{\texttt{ZX-Parse}}                           \\ \cmidrule(r){3-4}  \cmidrule(r){5-6}  \cmidrule(r){7-8}
\multicolumn{2}{l}{Match(\%)}  & \multicolumn{1}{l}{MCWQ-R} & \multicolumn{1}{l}{MCWQ} & \multicolumn{1}{l}{MCWQ-R} & \multicolumn{1}{l}{MCWQ} & \multicolumn{1}{l}{MCWQ-R} & \multicolumn{1}{l}{MCWQ}  \\ 
\midrule
\multicolumn{8}{l}{\textit{\textbf{Within-language}} (Supplement to Table \ref{monolingual}).} \\
\midrule
\multirow{3}{*}{MCD$_{1}$}             & EN                               & \multicolumn{2}{c|}{77.6}                              & \multicolumn{2}{c|}{$ 75.4_{\pm0.7}$}  & \multicolumn{2}{c}{$ 35.8_{\pm4.4}$}         \\ 
                                       & JA                               & 75.7                       & 43.6                     & 78.4                       & 47.6   & - & -                   \\ 

                                       & ZH                               & 74.7                       & 52.8                     & 74.0                       & 48.1     & - & -                 \\ 
\midrule
\multirow{3}{*}{MCD$_{2}$}             & EN                               & \multicolumn{2}{c|}{13}                                & \multicolumn{2}{c|}{$ 35.9_{\pm0.7}$}   & \multicolumn{2}{c}{$ 13.1_{\pm3.4}$}                    \\ 
                                       & JA                               & 32.2                       & 18.1                     & 30.9                       & 18.5        & - & -              \\ 

                                       & ZH                               & 31.5                       & 21.1                     & 38.7                       & 34.3       & - & -       \\ 
\midrule
\multirow{3}{*}{MCD$_{3}$}             & EN                               & \multicolumn{2}{c|}{24.3}                              & \multicolumn{2}{c|}{$ 54.4_{\pm3.5}$}   & \multicolumn{2}{c}{$ 22.8_{\pm2.5}$}               \\ 
                                       & JA                               & 61.0                       & 30.8                     & 65.8                       & 32.7         & - & -      \\ 
                                       & ZH                               & 47.2                       & 34.9                     & 67.1                       & 48.3         & - & -       \\ 

\midrule
\multicolumn{8}{l}{\textit{\textbf{Cross-lingual}} (Supplement to Table \ref{crosslingual}).} \\
\midrule

\multirow{2}{*}{MCD$_{1}$}   
                                       & JA                                                   & 0.06                       & 0.15                     & $ 42.6_{\pm1.7}$             & $ 28.8_{\pm4.8}$           & $ 9.5_{\pm3.5}$              & $ 10.2_{\pm2.2}$            \\ 
                                       & ZH                                                   & 0.08                       & 0.08                     & $ 43.0_{\pm1.0}$             & $ 41.7_{\pm0.9}$           & $ 9.3_{\pm3.6}$              & $ 10.7_{\pm2.1}$            \\ 
\midrule
\multirow{2}{*}{MCD$_{2}$}  
                                       & JA                                                   & 0.07                       & 0.08                     & $ 24.5_{\pm1.6}$             & $ 18.8_{\pm0.9}$           & $ 5.0_{\pm1.0}$              & $ 5.1_{\pm1.2}$             \\ 
                                       & ZH                                                   & 0.08                       & 0.07                     & $ 27.0_{\pm1.2}$             & $ 28.0_{\pm2.2}$           & $ 5.3_{\pm1.7}$              & $ 5.5_{\pm1.1}$             \\ 
\midrule
\multirow{2}{*}{MCD$_{3}$} 
                                       & JA                                                   & 0.18                       & 0.20                     & $ 39.0_{\pm2.9}$             & $ 26.2_{\pm2.8}$           & $ 11.7_{\pm0.8}$             & $ 10.2_{\pm1.3}$            \\ 
                                       & ZH                                                   & 0.20                       & 0.40                     & $ 43.2_{\pm3.2}$             & $ 35.2_{\pm3.6}$           & $ 13.4_{\pm0.7}$             & $ 11.1_{\pm1.8}$            \\
\bottomrule
    \end{tabular}}
    \caption{\label{tab:mcds}
    Detailed experiment results breakdown of 3 MCD splits, as supplement to Table \ref{crosslingual} and \ref{monolingual}.}
\end{table*}

\subsection{mT5$^{*}$}\label{appendix:additional}
In additional experiments, we freeze the mT5 encoders and train randomly initialized layers as mBART50$^{*}$ on English. The cross-lingual generalization results are shown in Table \ref{tab:additional}. While training decoder from scratch seemingly slightly ease cross-lingual transfer as also stated by \citet{sherborne-lapata-2022-zero}, the monolingual performance of mT5-small drops without pre-trained decoder. The results of mT5-large is consistent with \citet{qiu2022evaluating} which shows that increasing model size brings moderate improvement. However, the performance is still not comparable with mBART50$^{*}$, indicating that training paradigm does not fully account for the performance gap in Table \ref{crosslingual}.

While mT5 still struggle in zero-shot generation, the systematic hallucinations of country of origin mentioned in \S\ref{XL-analysis} disappear in this setup, due to the absence of pre-trained decoders which potentially encode the language bias.

\begin{table}[H]
    \centering
    \resizebox{0.48\textwidth}{!}{
\begin{tabular}{cc||cc|cc} 
\toprule
\multicolumn{2}{l}{Exact} & \multicolumn{2}{c}{\texttt{mT5-small$^{*}$}}                         & \multicolumn{2}{c}{\texttt{mT5-large$^{*}$}}                      \\ 
\cmidrule(r){3-4}  \cmidrule(r){5-6} 
\multicolumn{2}{l}{Match(\%)}   & \multicolumn{1}{l}{MCWQ-R} & \multicolumn{1}{l}{MCWQ} & \multicolumn{1}{l}{MCWQ-R} & \multicolumn{1}{l}{MCWQ}  \\ 
\midrule
\multirow{3}{*}{\transbox{MCD$_{\textrm{mean}}$}} & \textit{EN}                      & \multicolumn{2}{c|}{25.9}                              & \multicolumn{2}{c}{28.0}                               \\ 

                                       & JA                                                   & 1.0                        & 1.1                      & 4.0                        & 3.6                       \\ 

                                       & ZH                                                   & 1.2                        & 1.0                      & 4.2                        & 2.7                       \\ 
\midrule
\multirow{3}{*}{\transbox{Random}}                & \textit{EN}                                          & \multicolumn{2}{c|}{96.3}                              & \multicolumn{2}{c}{97.3}                               \\ 
                                       & JA                                                   & 6.3                        & 4.3                      & 11.3                       & 6.7                       \\ 

                                       & ZH                                                   & 5.5                        & 4.9                      & 13.7                       & 10.6                      \\
\bottomrule
\end{tabular}
    }
    \caption{Additional experiment results by replacing mBART50 with mT5 encoders: superscript ${^{*}}$ refers to the training paradigm of freezing pre-trained encoder as embedding layer and training randomly initialized encoder-decoder.}
    \label{tab:additional}
\end{table}


\section{Supplementary Analysis}\label{appendix:analysis}
\subsection{Structural Simplification}\label{appendix:simplification}
The train-test overlaps intuitively reflect the structural simplification, we show the numbers by structural cases and concrete examples in Table \ref{train-test}.

\begin{table*}[htb]
\centering
\small
\renewcommand{\arraystretch}{0.5}
\begin{tabular}{@{}ccccc@{}}
\toprule
    & \multicolumn{2}{c}{\textbf{EN}} & \textbf{JA} & \textbf{ZH} \\ \cmidrule(r){2-3} \cmidrule(r){4-4} \cmidrule(r){5-5} 
Possessive Case (Train/Test) & 0 / 49 & 49 / 0 & 49 / 49 & 27 / 27\\ \midrule
\small{SPARQL} & \multicolumn{4}{l}{\small{\texttt{( ?x0 ( wdt:P40|wdt:P355 ) ( ?x1 ) ) . ( ?x1 ( wdt:P106 ) ( wd:Q33999 ) )}}}
\\
ParseTree & 
\begin{minipage}[b]{0.3\columnwidth}
		\centering
		\raisebox{-.5\height}{\includegraphics[width=\linewidth]{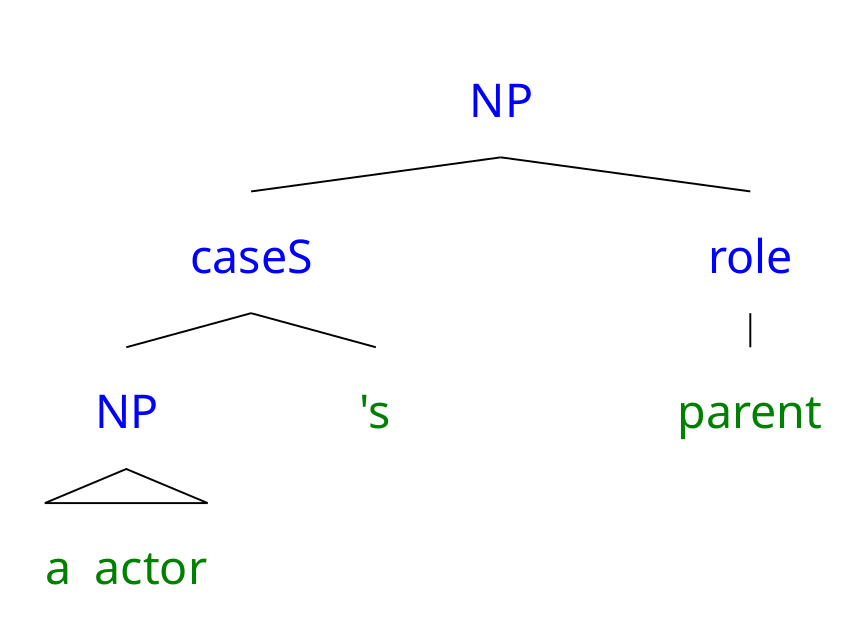}}
	\end{minipage}
 & 
\begin{minipage}[b]{0.3\columnwidth}
		\centering
		\raisebox{-.5\height}{\includegraphics[width=\linewidth]{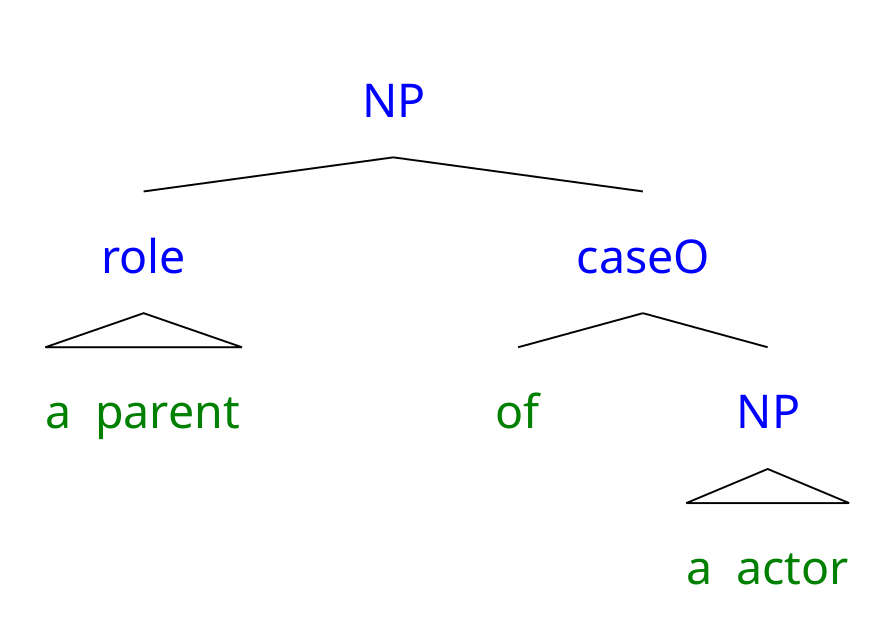}}
	\end{minipage}
&  
\begin{minipage}[b]{0.3\columnwidth}
		\centering
		\raisebox{-.5\height}{\includegraphics[width=\linewidth]{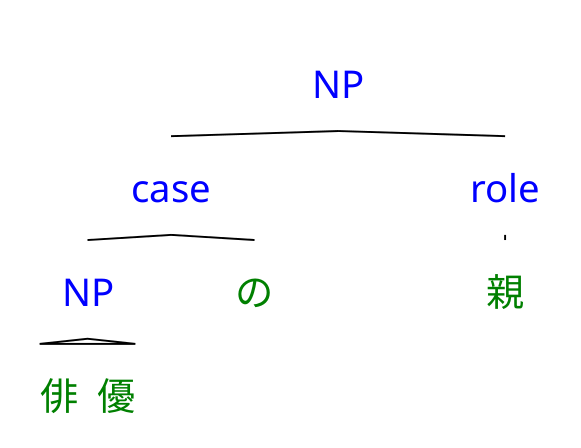}}
	\end{minipage}
& 
\begin{minipage}[b]{0.3\columnwidth}
		\centering
		\raisebox{-.5\height}{\includegraphics[width=\linewidth]{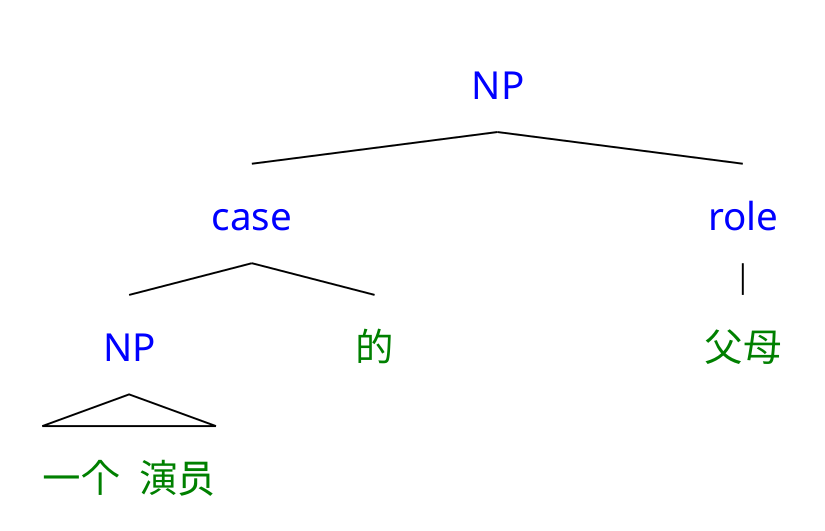}}
	\end{minipage}
\\
\midrule 
Preposition in Passive & 0 / 7 & 7 / 0 & 7 / 7 & 7 / 7   \\ \midrule
\small{SPARQL} & \multicolumn{4}{l}{\small{\texttt{(( ?x0 ( wdt:P750 , wdt:P162|wdt:P272 ) ( ?x1 ) )}}} \\ 
ParseTree & 
\begin{minipage}[b]{0.3\columnwidth}
		\centering
		\raisebox{-.5\height}{\includegraphics[width=\linewidth]{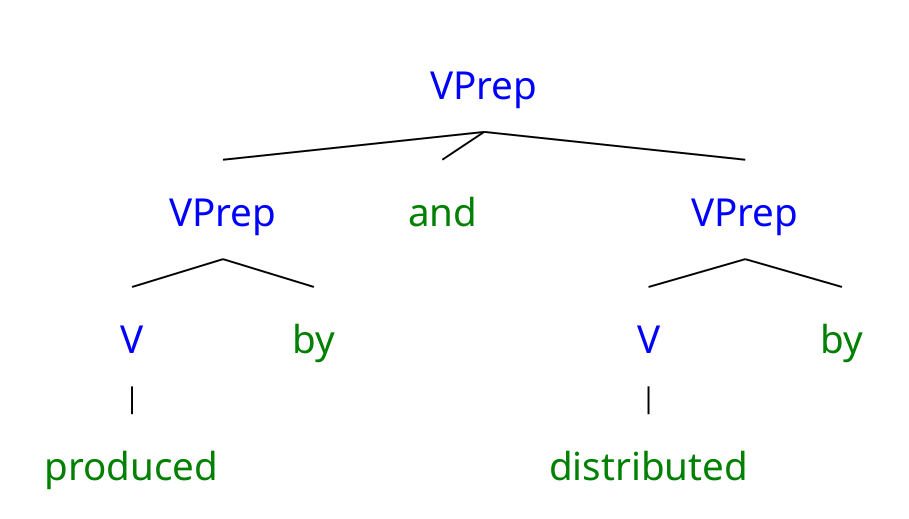}}
	\end{minipage}
 &
\begin{minipage}[b]{0.3\columnwidth}
		\centering
		\raisebox{-.5\height}{\includegraphics[width=\linewidth]{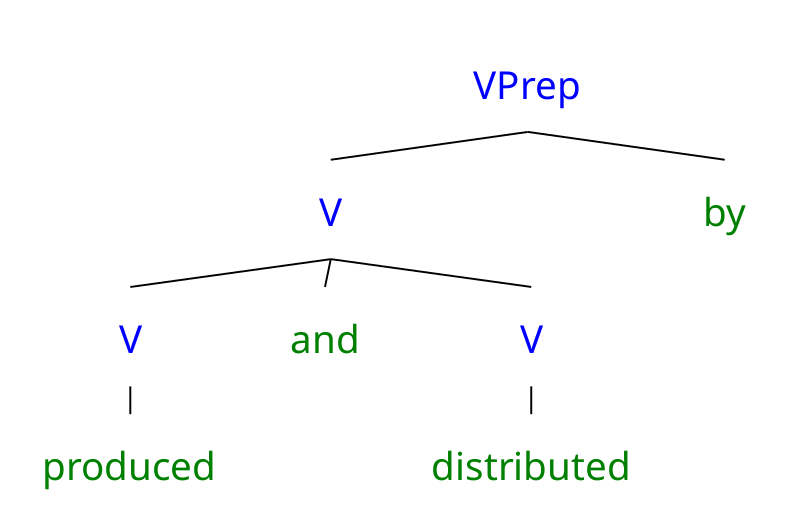}}
	\end{minipage}
&  
\begin{minipage}[b]{0.3\columnwidth}
		\centering
		\raisebox{-.5\height}{\includegraphics[width=\linewidth]{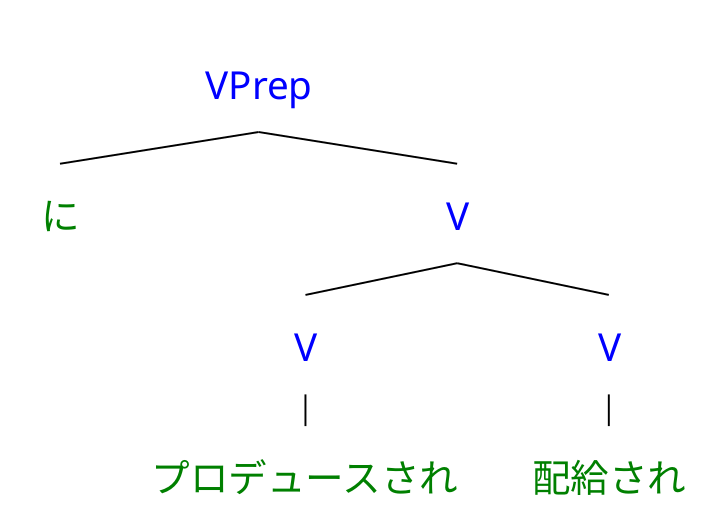}}
	\end{minipage}
& 
\begin{minipage}[b]{0.3\columnwidth}
		\centering
		\raisebox{-.5\height}{\includegraphics[width=\linewidth]{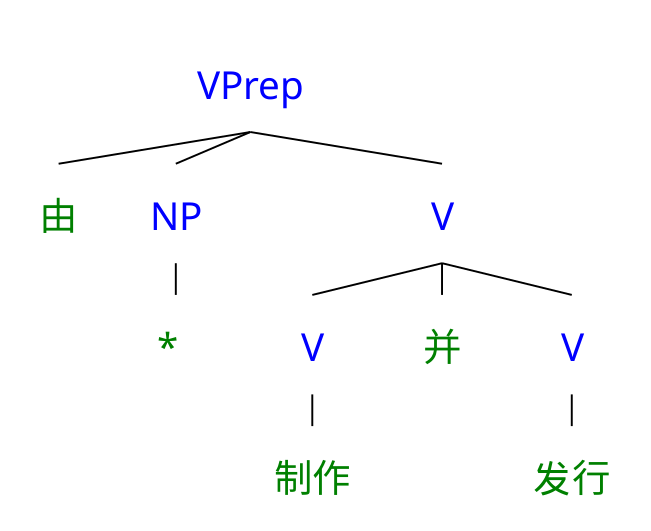}}
	\end{minipage}
\\
\midrule 
Interrogative Pronoun & 0 / 4 & 4 / 0 & 2 / 2 & 4 / 4   \\ \midrule
\small{SPARQL} & \multicolumn{4}{l}{\small{\texttt{SELECT DISTINCT ?x0 WHERE lb ( ?x0 ( wdt:P106 ) ( wd:Q36834 ) ) .}}} \\ 
ParseTree & 
\begin{minipage}[b]{0.3\columnwidth}
		\centering
		\raisebox{-.5\height}{\includegraphics[width=0.7\linewidth]{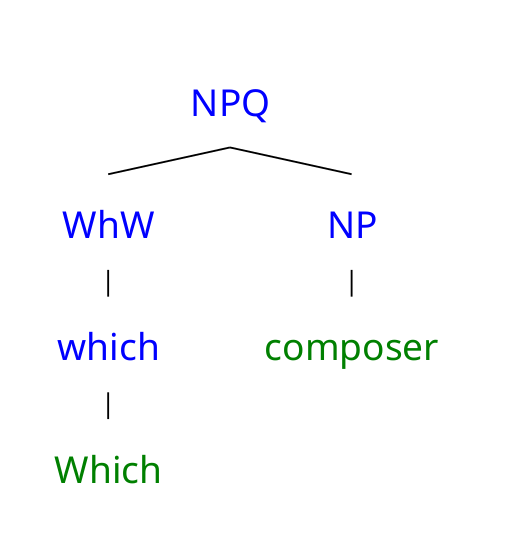}}
	\end{minipage}
 &
\begin{minipage}[b]{0.3\columnwidth}
		\centering
		\raisebox{-.5\height}{\includegraphics[width=0.7\linewidth]{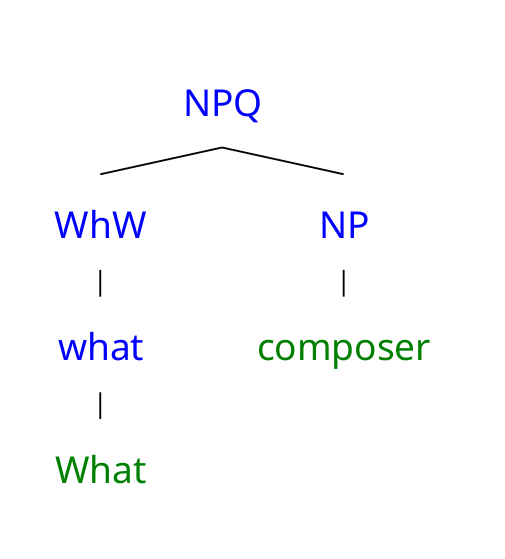}}
	\end{minipage}
&  
\begin{minipage}[b]{0.3\columnwidth}
		\centering
		\raisebox{-.5\height}{\includegraphics[width=0.7\linewidth]{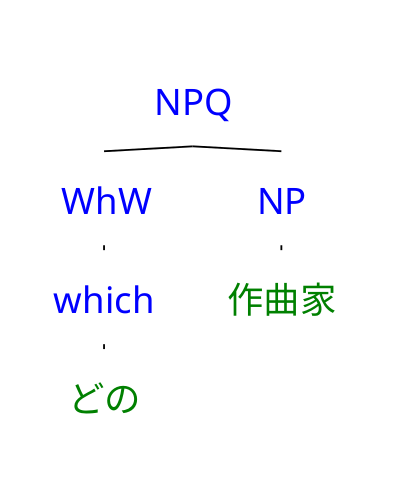}}
	\end{minipage}
& 
\begin{minipage}[b]{0.3\columnwidth}
		\centering
		\raisebox{-.5\height}{\includegraphics[width=0.7\linewidth]{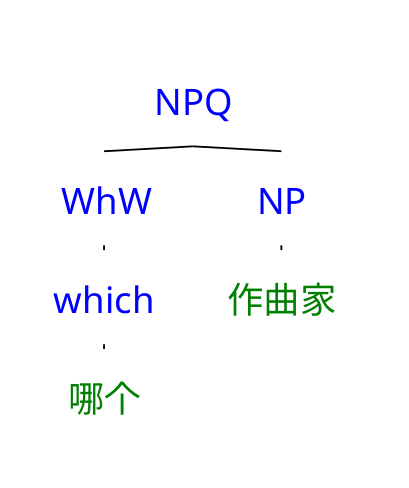}}
	\end{minipage}

\\ \bottomrule
\end{tabular}
\caption{\label{train-test}
The train-test overlap of 3 MCD splits for JA and ZH together with their EN source patterns. We present the union of the 3 intersections, from which we observe 3 types of structures (part-of-speech tags in blue) leading to structural ``simplification''. We provide concrete examples (green) of the structures and their \textit{common} SPARQL fragments. EN possesses multiple structures for each of the fragments, while JA and ZH possess only one (considering the specific context).
}
\end{table*}

\subsection{Representation Alignment in ZX-Parse}\label{appendix:align}
\begin{figure}[h]
	\centering \includegraphics[width=0.48\textwidth]{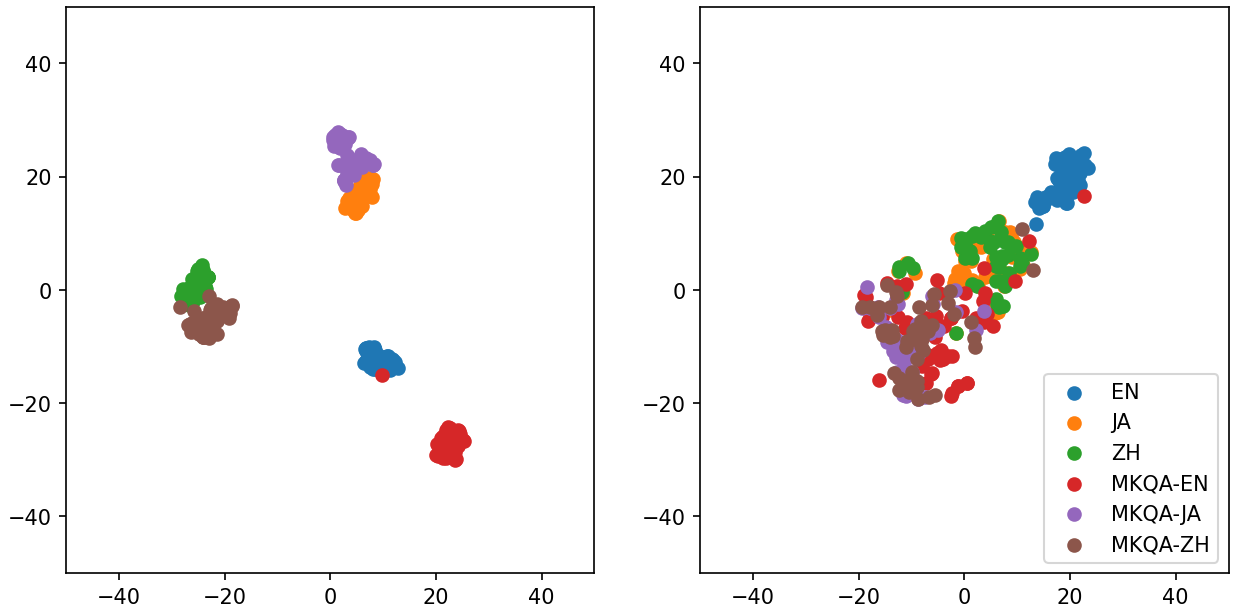}
\caption{\label{alignment}
t-SNE analysis on \textbf{MCWQ-R} and MKQA samples. We show the latent representations (of 50 samples per category) by mBART50 embedding layers (left)
and by the ZX-Parse encoder (right) i.e., before and after the aligning layer trained with MKQA bi-text.
}
\end{figure}

We analyze the representations before and after the trained aligning layer with t-SNE visualization as \citet{sherborne-lapata-2022-zero} do. 
Figure \ref{alignment} illustrates an example, the representations of compositional utterances (especially English) are distinct from natural utterances from MKQA, even after alignment, which demonstrates the domain gap between the 2 categories of data. Nonetheless, the mechanism performs as intended to align representations across languages.
 
\subsection{Accuracy by Complexity}\label{appendix:acc.by.complexity}
\begin{figure}[tb]
    \centering
	\includegraphics[width=0.45\textwidth]{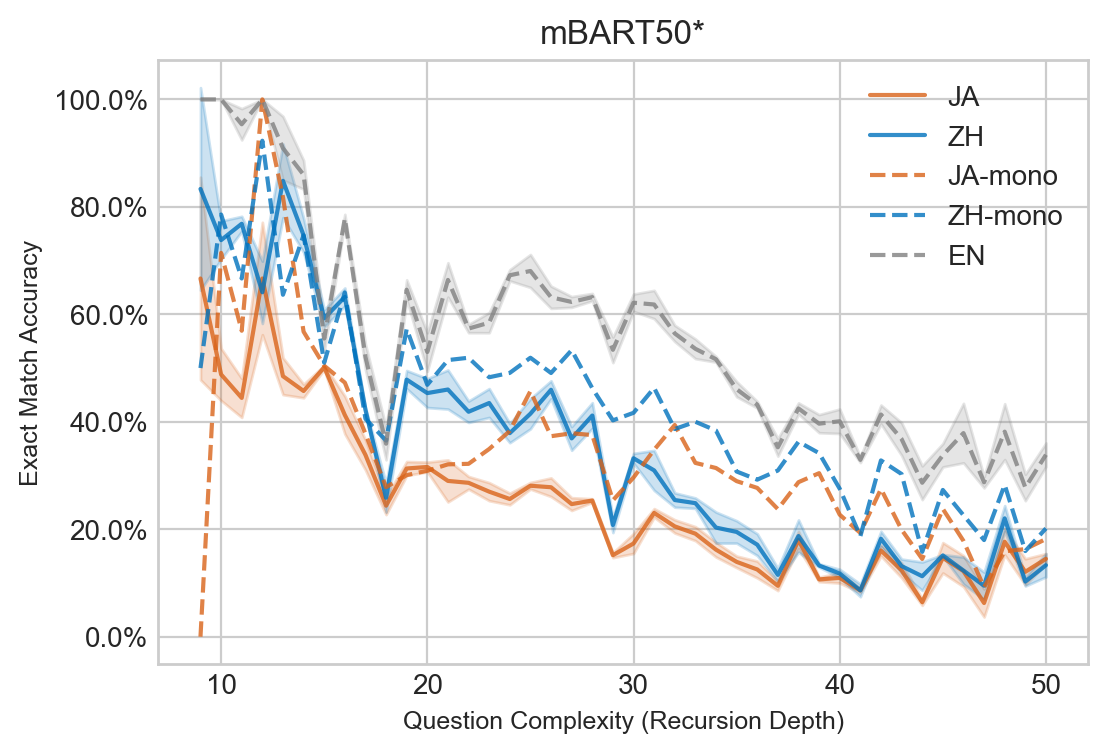} \\
    \caption{Accuracy of mBART50$^{*}$ on \textbf{MCWQ} varies against increasing question complexity, averaged over 3 MCD splits. Dashed lines refer to within-language generalization performance.}
    \label{fig:acc.vs.cplx.MCWQ.mBART}
\end{figure}

We present the accuracy by complexity on MCWQ in Figure 
\ref{fig:acc.vs.cplx.MCWQ.mBART}. We notice the gaps between monolingual and cross-lingual generalization are generally smaller than on MCWQ-R (see Figure \ref{acc.vs.cplx}). This is ascribed to the systematicity of GT errors---such (partially) systematical errors are fitted by models in monolingual training, and thus cause falsely higher performance on the test samples possessing similar errors.

\begin{figure}[htb]
    \centering
	\includegraphics[width=0.45\textwidth]{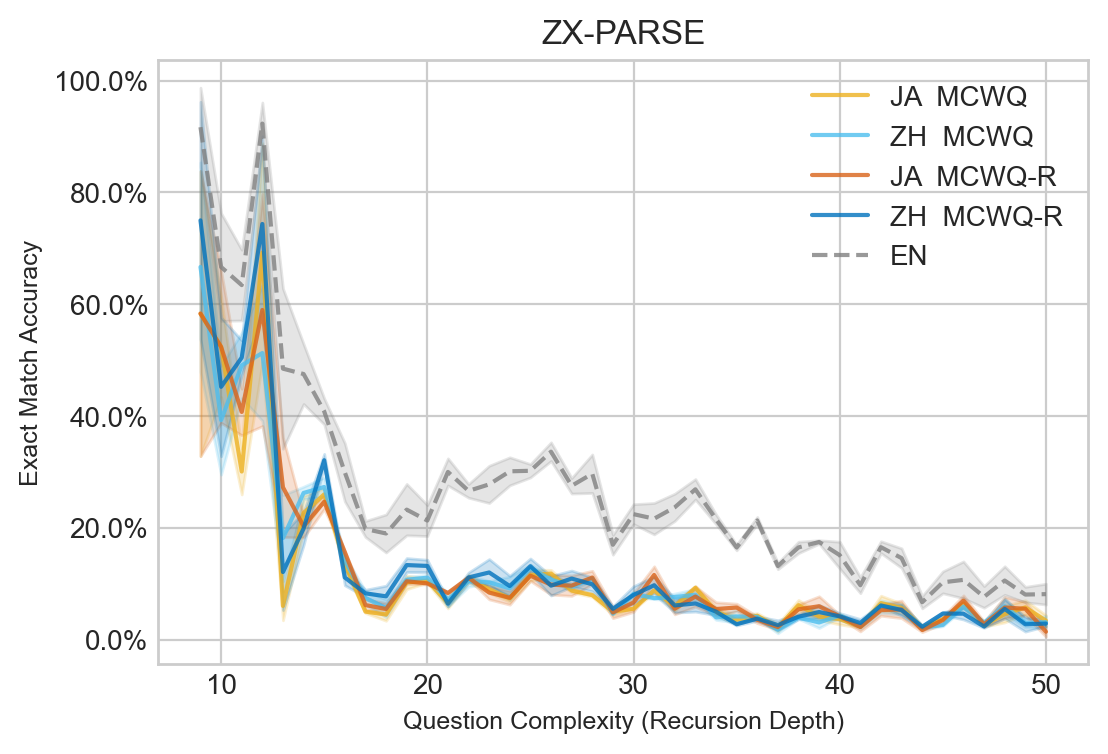} \\
    \caption{Cross-lingual generalization accuracy of ZX-Parse on both datasets varies against increasing question complexity, averaged over 3 MCD splits. EN monolingual results are presented in the dashed line.}
    \label{fig:acc.vs.cplx.ZXP}
\end{figure}

Figure \ref{fig:acc.vs.cplx.ZXP} shows the cross-lingual results of ZX-Parse on both datasets. While the accuracies are averagely lowered, the curves appear to be more aligned due to the mechanism.

\end{document}